\documentclass[sigconf]{acmart}

\usepackage{booktabs} 

\setcopyright{none}

\begin{document}
\title{GA-SVM for Evaluating Heroin Consumption Risk}

\author{Sean-Kelly Palicki}
\affiliation{%
  \institution{Technical University of Munich}
  \city{Munich} 
  \state{Germany} 
}
\email{sean.palicki@tum.de}

\author{R. Muhammad Atif Azad}
\affiliation{%
  \institution{Birmingham City University}
  \city{Birmingham} 
  \state{England} 
}
\email{atif.azad@bcu.ac.uk}

\begin{abstract}
There were over 70,000 drug overdose deaths in the USA in 2017. Almost half of those involved the use of Opioids such as Heroin. This research supports efforts to combat the Opioid Epidemic by further understanding factors that lead to Heroin consumption. Previous research has debated the cause of Heroin addiction, with some explaining the phenomenon as a transition from prescription Opioids, and others pointing to various psycho-social factors. This research used self-reported information about personality, demographics and drug consumption behavior to predict Heroin consumption. By applying a Support Vector Machine algorithm optimized with a Genetic Algorithm (GA-SVM Hybrid) to simultaneously identify predictive features and model parameters, this research produced several models that were more accurate in predicting Heroin use than those produced in previous studies. Although all factors had predictive power, these results showed that consumption of other drugs (both prescription and illicit) were stronger predictors of Heroin use than psycho-social factors. The use of prescription drugs as a strong predictor of Heroin use is an important though disturbing discovery but that can help combat Heroin use. 
\end{abstract}

%
%



\maketitle

\section{Introduction}
\label{sec:intro}
 Drug addiction is a major public health problem and the prevalence of drug use has been increasing globally \cite{Aptekin18}. Drug addiction is a disorder that includes the use of highly addictive substances despite the adverse effects on the individual and society \cite{Ali11}. The United Nations Office of Drugs and Crime estimated that one out of every 20 people between the ages of 15 and 64 is addicted to 
 illicit drugs that amounts to a total of 246 million people worldwide in 2013 \cite{UN16}. In 2017, there were a total of 70,237 drug overdose deaths in the USA alone \cite{Scholl19}. 

However, collecting information on illicit drug use is difficult and this can make service delivery difficult for healthcare providers such as those dealing with mental health \cite{Garg15, Stacey85}. Anonymized web-based surveys give estimates of general use patterns; however, they typically under-represent the actual extent of drug use, and their validity cannot be guaranteed for individuals in all circumstances \cite{Garg15}. As it can be difficult to collect accurate information regarding an individual's drug consumption, predictive models are needed to help identify individuals at risk.  

Previous research indicates some factors leading to drug consumption. These include personality, socio-economic status and risky behavior, as well as genetic and environmental factors \cite{Walt04, Habibi13, Zilb18, Ali11}. One study found that although some predictors such as personality traits of impulsivity and neuroticism were common among various addicted populations, the predictors for {\em different types} of addictions differed significantly \cite{Zilb18}. Their results show that more work is still needed to identify the causes of specific addictions. 

Data mining and artificial intelligence have been applied to understand distinct drug consumption patterns \cite{Fehr15, Wu18}. The authors of one such study compared several classifiers including kNN, Decision Tree, Random Forest, and Naive Bayes to determine the relationship between personality, demographics factors, and consumption of 18 legal as well as illicit drugs \cite{Fehr15}. They found that the best subset of input features, and classifier type varied across the different drugs. Notably for this study, they also found significant clusters for consuming several drugs. The researchers merged drugs with highly correlated consumption into modules called 'Pleiades of drugs'. These clusters included the Heroin pleiad (crack, cocaine, methadone, heroin), Ecstacy pleiad (amphetamines, cannabis, cocaine, ketamine, LSD, mushrooms, legal highs, ecstacy) and the Benzodiazepines pleiad (methadone, amphetamines, and cocaine). Their results showed that consumption of some drugs are easier to predict than others, and that there may exist relationships in drug consumption behaviors which could be used to improve predictions. 
 
 Building upon previous work, this study found the optimal subset of personality factors, socio-demographic factors, and previous drug consumption behaviors, to more accurately predict Heroin consumption. This paper begins by introducing the reader to the problem of Heroin addiction in section~\ref{sec:background}; section~\ref{sec:problem} describes the data source, and the fitness function employed; section
\ref{sec:methods} discusses the optimization and classification algorithms, and how they can be employed in tandem in this study; section~\ref{sec:methodology} details data pre-processing as well as implementation of the modelling methodology; section \ref{sec:experiments1} presents the experimental details, the statistical significance of the achieved results and a discussion of how they impact the on going debate in the field of Heroin addiction and its causes; finally, the paper concludes in section~\ref{sec:conclusions}.

\section{Causes of Heroin Addiction}
\label{sec:background}
Substance abuse is an area of research that concerns practitioners of both physical and mental health as drug use is correlated with numerous negative health outcomes \cite{Horgan01}. Within this field of research, the predictability of heroin use is particularly important, as heroin is highly accessible and deadly. There are two major theories for the onset of heroin addiction, the first holds prescription medication responsible, whereas the second describes social and demographic factors \cite{Dasgupta18}.  

Heroin is included in a highly addictive class of drugs known as Opioids. Opioids are unique in that, although highly addictive and deadly, they can be legally obtained in many countries through prescription pain medications such as OxyContin and Fentanyl. Although legal drug prescriptions containing Opioids account for two-thirds of all deaths due to drug overdose and relate to increases in consumption of the illegal drug Heroin, they are increasingly prescribed \cite{Guy17}. To put this into perceivable terms, out of a total of 70,237 drug overdose deaths in the USA in 2017, 17,029 involved prescription Opioids and another 15,482 were due to the illicit, non-presciption, Opioid drug known as Heroin \cite{Scholl19}.

measuring the predictive power of personality and demographic factors on a variety of drugs including Heroin, \cite{Fehr15} found that Age, Impulsivity, and Gender were moderately accurate predictors of Heroin use. Using the same data from their previous research, this study found that including knowledge about consumption of other drugs, such as Cocaine and Amphetamines, produced models of Heroin consumption which outperformed those with personality and demographic factors alone. Furthemore, we found that factors related to drug consumption behavior were more significant than other factors such as Gender and Age. 

\section{Problem Instance}
\label{sec:problem}

This study used the Drug Consumption Data Set available through the UCI Machine Learning Repository. The data set has been prepared for use with classification algorithms and contains a set of 1,885 individuals measured across 32 attributes, including personality type, demographic factors, and drug consumption behavior. The data set was assembled between 03/2011 and 03/2012 through an online survey \cite{Fehr15}. \cite{Fehr15} used the data set to test whether personality type can predict drug use. Across different classifiers and drug categories, they achieved an average classification accuracy of 70\%.

Using the results of \cite{Fehr15} as a benchmark, this study sets out to build accurate predictive models for Heroin consumption. However, improving upon the previous research, this study employs a meta-heuristic to simultaneously select optimal subsets of features as well as model parameters.

 The objective function (also called the {\em fitness function}) for this optimisation is similar to that in some previous work \cite{Huang06} that employed a Genetic Algorithm (GA) \cite{Holland75, Davis91, Goldberg89} to optimise the parameters and the set of predictive features for Support Vector Machines (SVM); hence the overall classification algorithm used here is a GA-SVM hybrid. For a set of features $F = f_1, f_2, f_3, ...f_n$ the objective is to find a subset $F' \subseteq F$ that maximizes $Fitness$ where

\begin{equation}
  \label{eq-fit}
  Fitness = W_A * Accuracy + W_F * (\sum_{i=1}^{n_f} F_i)^{-1}.
\end{equation}

\par\noindent Here $Accuracy$ is the product of specificity and sensitivity; 
$F_i$ is 1 when feature $i$ is selected and 0 otherwise; and $W_A$ and $W_F$ are relative weights for $Accuracy$ and the size of the feature set respectively. Note, the fitness function maximises accuracy while minimising the number of features used by the classification model produced by the SVM. A feature-minimal yet accurate model is desirable. 

 Note, the measurement of classification accuracy is an open question. Some approaches popular in research, such as Kappa Index Agreement (KIA), have been shown to misrepresent accuracy \cite{Pont11}. Hence, this study chose a straightforward approach to define accuracy as the product of specificity and sensitivity \cite{Kohavi97, Guyon03}.
 
 Note, while it is possible to use multi objective optimisation due to the twin objectives in equation (\ref{eq-fit}), the simple linear combination of the objectives as above  yielded satisfactory results in this paper. As such the purpose of this paper is not to conduct an exhaustive comparison of various optimisation approaches; instead, the desire is to obtain satisfactory results for this interesting and impactful application of Machine Learning. However, further work can also employ other optimisation techniques for comparative purposes. 
 
\section{A Hybrid Approach to Classification}
\label{sec:methods}
This section briefly introduces the two components of the adopted classification approach, that is, Genetic Algorithms and Support Vector Machines and then reviews some past precedent in hybridizing the two methods together to produce accurate predictive models.

\subsection{Support Vector Machines}
Support Vector Machines (SVMs) are used widely in healthcare classification problems \cite{Lin08, Xu18}. By using a set of flexible parameters to create an optimal hyperplane between classes, they achieve accurate results. SVMs are also robust in that they may be used with many different types of data (binary, real-valued, integer, text). Importantly, they will work with the current data because they were effective on data sets with less than 2,000 observations \cite{Huang06, Huang07}. 

However, SVMs are sensitive to input features as well as to changes in parameters \cite{Keerthi03}. Accurate SVMs require proper parameter settings. This research used an SVM algorithm with a Radial Bases Function (RBF) kernal. The two main parameters of this pairing are $C$ to assign a penalty for error in the SVM algorithm, and gamma ($\gamma$) to determine the effect of error in the RBF function. Instead of using the default values for these two values, this paper optimises these parameters alongside the set of input features with a Genetic Algorithm that is described next. 

\subsection{Genetic Algorithm}
The theoretical and practical underpinnings of GA as a global search strategy to optimization through population based stochastic mechanisms were introduced in the 1970s  \cite{Holland75, Davis91, Goldberg89}. Genetic Algorithms (GAs) are in a class of computational approaches inspired by Darwin's theory of evolution by natural selection \cite{Holland75}. 

Darwin described natural selection as the process by which individuals possessing traits which allow them to adapt to environmental pressures tend to survive and reproduce in the greatest numbers \cite{Darwin1882}. Similarly, GAs start with a population of candidate solutions to an optimization problem (individuals), of which, the most fit are bred to create additional generations of fitter individuals. The process continues until the population converges on an optimal solution, or time runs out.

\subsection{Optimizing Model Parameters and Feature Subset using GA-SVM Hybrid}
GAs are used as a metaheuristic to achieve optimal SVMs for classification, using a method called GA-SVM Hybrid \cite{Ahmad14, Bab10, Tao18, Lin08}. GA-SVM Hybrid has been competitive against other SVM based hybrid algorithms such as PSO-SVM, NN-SVM, and ACO-SVM \cite{Ren10, Alba07, Prasad10}. In all cases, the optimization algorithm identifies a set of features and parameters to be included in a SVM classification model.  

\begin{figure}[htb]
\vskip 0.2in
\begin{center}
\centerline{\includegraphics[width=\columnwidth]{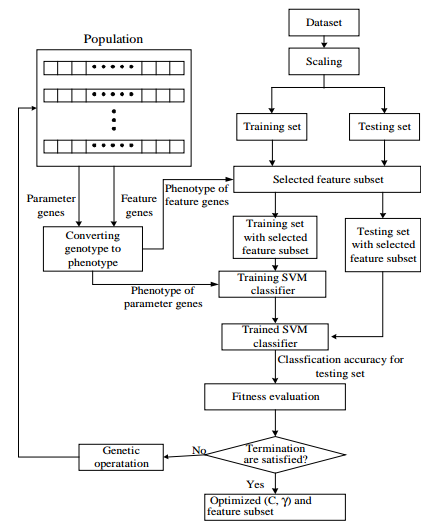}}
\caption{Architecture of SVM + GA with simultaneous feature and parameter optimization \cite{Huang06}.}
\label{fig:process}
\end{center}
\vskip -0.2in
\end{figure}

Feature selection is essential to classification accuracy, efficient computation, and interpretability of a machine learning model \cite{Vip14, Guyon03}. There are multiple approaches to feature selection. The approach used in this research (Figure \ref{fig:process}) trained an SVM while simultaneously selecting optimal features and parameters \cite{Huang06, Kohavi97}.

In GA based feature selection and parameter tuning, each individual is a fixed length string (chromosome) corresponding to a set of potential features and SVM parameters. In this research each chromosome had 32 real numbers, representing 30 features from the data set and two SVM parameters associated with an RBF kernal ($C$ and $\gamma$) as discussed before;  together the resulting chromosome is evaluated with the fitness function (Figure \ref{fig:process}). Section \ref{sec:gasvm} further justifies the composition of chromosome and its translation into a solution. 
 

\begin{figure}[htb]
\vskip 0.2in
\begin{center}
\centerline{\includegraphics[width=\columnwidth]{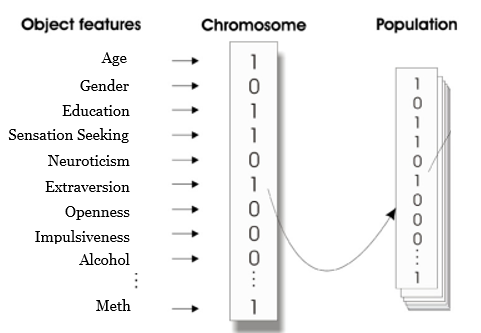}}
\caption{Chromosome of GA for feature selection \cite{Stein05}.}
\label{fig:mapping}
\end{center}
\vskip -0.2in
\end{figure}


\section{Methodology}
\label{sec:methodology}
The following section details the research design and techniques used in this study.

\subsection{Data Pre-Processing}
\label{sec:datapreprocessing}
The target variable of this study was heroin consumption. The time period for which to define heroin consumption is an important consideration since there will be a different population who had consumed Heroin within the last week than those who had within the last year, decade, or in their lifetime. This problem is discussed more in \cite{Fehr15}. For this research, drug consumption is defined using a `Decade-Based' Binary Classification scheme in which `Non-Users' were those had either `Never Used' or `Used Longer than a Decade'. Whereas those who had used within the last decade were classified as `Users'. Table \ref{tab:users} shows the distribution of drug consumption using this cutoff, consumption for all drugs is shown.

Since Heroin users made-up only 11.25\% of the recorded data instances (see Table \ref{tab:users}), the binary User/Non-User classification problem was unbalanced. SVM can work with unbalanced data; however, the training and test data needed to accurately reflect distributions in the overall data set. As such, k-Fold Cross Validation, which uses a method similar to stratified sampling, was used to make training and test data sets. It has been shown that 3-Fold Cross Validation can be effective when using SVM with an RBF Kernal for classification \cite{Anguita12}. This research uses 3-Fold Cross Validation to create unique splits of data with well-represented training (60\%) and test (30\%) data.

\begin{table}[htb]
\centering
\caption{\label{tab:users} Number and Percentage of Drug Users.}
\small
\begin{tabular}{ccc}
\hline
Drug & Number & Percent User \\
\hline
Alcohol & 1817 & 96.39\%  \\ 
Amphetamines & 679 & 36.02\% \\ 
Amyl Nitrite & 370  & 19.63\% \\ 
Benzodiazepines & 769  & 40.80\% \\ 
Cannabis & 1265 & 67.11\%  \\ 
Chocolate  & 1850 & 98.14\%  \\ 
Cocaine & 687 & 36.45\% \\ 
Caffeine & 1848 & 98.04\% \\ 
Crack & 191 & 10.13\% \\ 
Ecstasy & 751 & 39.84\% \\ 
Heroin* & 212 & 11.25\% \\ 
Ketamine & 350 & 18.57\% \\ 
Legal Highs & 762 & 40.42\% \\ 
LSD & 557 & 29.55\% \\ 
Methadone & 417 & 22.12\% \\ 
MMushrooms & 694 & 36.82\% \\ 
Nicotine & 1264 & 67.06\% \\ 
VSA & 230 & 12.20\% \\ 
\hline
\\
\end{tabular}
\end{table}

\begin{table}[htb]
\centering
\caption{\label{tab:convert} Translating Genotype to Phenotype.}
\begin{tabular}{|l|l||l|l||l|l|l|} \hline
$F_{1}$ & $F_{2}$ & $F_{3}$ & ... & $F_{n}$ & $C$ & $\gamma$ \\ \hline
0.8 & 0.2 & 0.6 & ... & $F_{n}$ & 42 & 2.33 \\ \hline
1 & 0 & 1 & ... & $F_{n}$ & 42 & 2.33 \\ \hline
\end{tabular}
\end{table}

\subsection{ GA-SVM Hybrid Model in R}
\label{sec:gasvm}
As explained before, to explore the search space of possible SVM paramaters while simultaneously finding the best possible feature combination, the chromosomes for the GA included both the set of features as well as SVM Gamma and Cost parameters. Because $C$ (Cost parameter) and $\gamma$ are real values, the entire chromosome comprises of real numbers. Therefore, the real numbers corresponding to the features were rounded to 1 or 0, with a 0.5 cutoff. A `1' meant that the corresponding feature would be included into SVM classification model, and a `0' meant otherwise; see Table \ref{tab:convert}. As suggested by  $C$ and $\gamma$ ranged between 1-100, and 1-10 respectively. The values of $C$ and $\gamma$ were expected to differ with the changes in input variable. Only chromosomes with at least 1 feature were included when training the SVM classifier. Furthermore, chromosomes which selected 0 features were given the worst possible fitness.  

The predicted values from the SVM classifier were compared against the testing set; Sensitivity, Specificity, and number of Features were calculated to give the fitness function values, which were used to evaluate the quality of each model. The evolutionary process continued until the population converged on the best solution (20 generations without producing a better solution) or 100 generations had been produced. This was repeated 30 times for each of the 3 folds of data, resulting in 90 iterations for each experimental condition.

The theoretical overview (Figure \ref{fig:process}) was implemented in R using a combination of the GA and e1071 packages. Calculations were made in parallel over 4 cores with the foreach package on an Intel Core i7-8705G CPU.

\section{Experiments}
\label{sec:experiments1}
This study compared the effects of changes in fitness function weights on the classification accuracy and feature reduction of the GA-SVM Hybrid algorithm. Experiments are run using 3-fold cross validation, and average results are reported. 

\subsection{Dataset Description}
\label{sec:dataset}
There were a total of 1,885  observations in the dataset. Included were demographic variables, personality characteristics, and consumption behavior for prior drug use. Of the total sample, there were a near equal number of males and females (943/942), primarily from the UK and USA (55.4\% and 29.5\% respectively); 
60\% of the sample were between 18 and 34 years old, and the rest 35 years and older.
The sample is largely white (91\%), English speaking (93.7\%), and have completed a degree or professional certificate (59.5\%).

\begin{figure}[htb]
\vskip 0.2in
\begin{center}
\textbf{Impact of Mutation and Crossover Rate}\par\medskip
\centerline{\includegraphics[height=7cm,width=1\columnwidth]{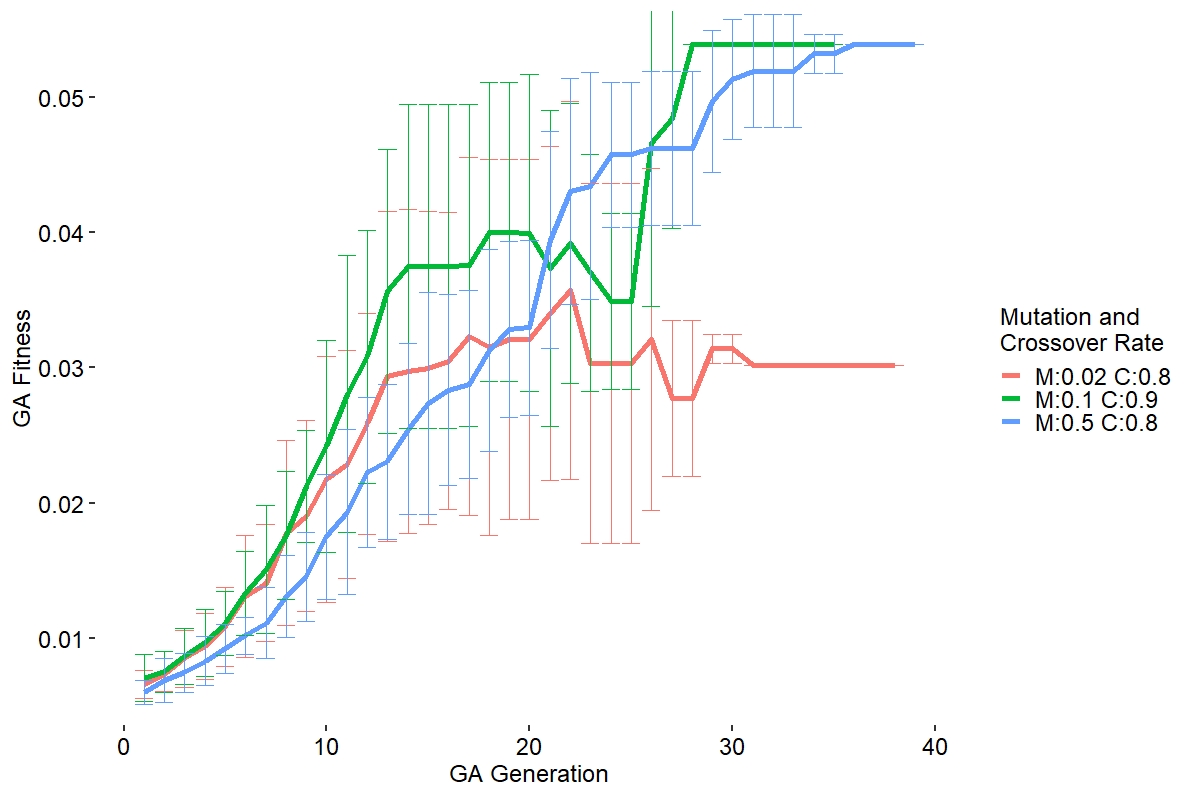}}
\caption{Mutation and Crossover Rate resulted in different patterns of learning.}
\label{fig:comparison}
\end{center}
\vskip -0.2in
\end{figure}

\subsection{Parameter Selection}
\label{sec:parameters}
Major considerations for running a GA include parameter settings of mutation rate, crossover rate, level of elitism, population size, and number of generations. When using a GA-SVM hybrid, researchers have disagreed in their GA settings. For example, population size in some research is as little as 10 individuals while in others it is up to 500 \cite{Tan09, Huang06}. Based on initial experiments and computational considerations, this study used a population of 100, and elitism of 10\%. Another aspect, important for the breeding of new solutions, are the mutation and crossover rates. For some exploratory experiments, we tested different levels of mutation and crossover seen in other GA-SVM Hybrid implementations to decide on the settings for the full experiment \cite{Oztekin18, Kharrat10, Huang06}. The results showed that mutation rate of 2\% and crossover rate of 80\% were not a good setting for this research (Figure \ref{fig:comparison}); instead, some higher mutation rates of 10\% or more were better as they produced high fitness, high accuracy, and a quick learning rate. Therefore, Table \ref{tab:ga-params} summarises the parameters used for the remaining experiments in this paper; it uses a mutation rate of 10\% and crossover rate of 90\% as this combination performed well in initial experimentation. 

\begin{table}[htb]
\centering
\caption{\label{tab:ga-params} GA Parameters.}
\small
\begin{tabular}{cc}
\hline
Parameter & Value \\
\hline
Population Size & 100  \\ 
Maximum Gens. & 100  \\ 
Total Iterations & 30  \\ 
Cross-Validation & 3-fold  \\ 
Crossover ($P$) & 0.9  \\ 
Mutation ($P$) & 0.1  \\ 
Elitism & 10  \\ 
Convergence & 20-run  \\ 
\hline
\\
\end{tabular}
\end{table}

\subsection{Experimental Conditions}
\label{sec:experiments}
Though this it is a common approach to use a fitness function with weights on Accuracy and Number of Features, the setting of weights is left up to discretion of the researcher. There is no consensus on the proper settings for these weights, and researchers choose different values without always explaining their choice, or comparing to alternatives \cite{Oztekin18, Kharrat10, Huang06, Tan09, Alba07}. 

The fitness function is the most important part of the GA algorithm because this is used to assess the quality of potential solutions. As GA and SVM parameters affect the outcome of the algorithm, so do weights in a weighted fitness function. Therefore, we compare the effect of setting relative weights for accuracy and number of features on the identification of good SVM parameters and features; Table \ref{tab:conditions} shows the various combinations of weights tried in this paper.

\begin{table}[htb]
\centering
\caption{\label{tab:conditions} Sets of weights for Accuracy and Number of Features tried in this paper.}
\small
\begin{tabular}{cccccc}
\hline
Fitness Component & 1 & 2 & 3 & 4 & 5 \\ 
\hline
Accuracy & 0.8 & 0.65 & 0.5 & 0.35 & 0.2 \\ 
Feature & 0.2 & 0.35 & 0.5 & 0.65 & 0.8 \\ 
\hline
\\
\end{tabular}
\end{table}

\subsection{Results}
\label{sec:results}

\begin{figure}[htb]
\vskip 0.2in
\begin{center}
\textbf{Fitness Growth Among Experimental Conditions}\par\medskip
\centerline{\includegraphics[height=7cm,width=1\columnwidth]{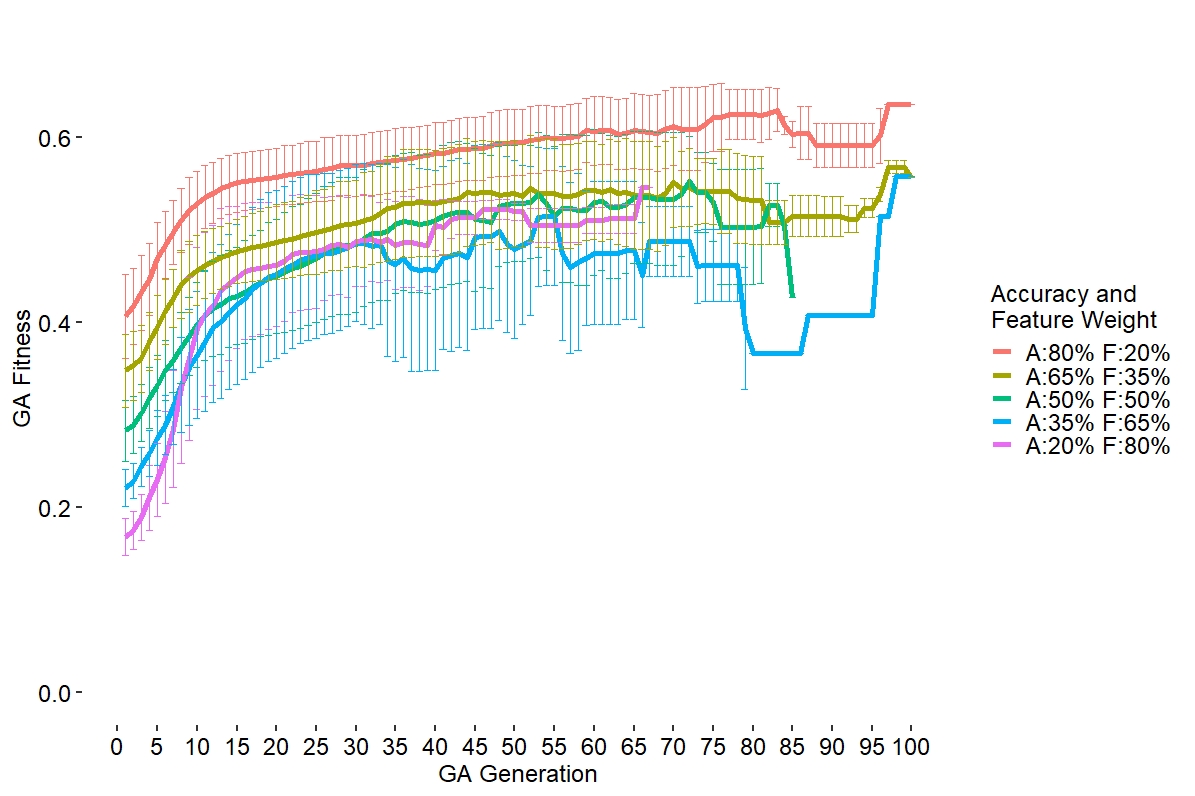}}
\caption{All conditions resulted in improved fitness over generations. The 80:20\% condition was the most fit.}
\label{fig:groups}
\end{center}
\vskip -0.2in
\end{figure}

As seen in Figure \ref{fig:groups}, all conditions improved over successive generations. As seen in the average with 95\% CI, there was variability in the rate of improvement, the maximum level of fitness, and the number of generations until convergence on an optimal solution. When Accuracy was set to 20\% and Number of Features to 80\% the rate of learning was the quickest, as the fitness grew rapidly in 15 generations, and made slow progress until final convergence in no more than 65 generations. When Accuracy was 35\% and Features 65\% there was a trend, over one of the folds, where the fitness drops after 75 generations, and remains with very little diversity in the population until a rapid growth after generation 95. Though all other conditions had similar fitness scores between generations 35 and 65, the 80:20\% condition was significantly more fit overall.

However, since fitness is computed differently in each setup, we also compare the accuracy and the number of features separately. Table~\ref{tab:fold_avg} shows the average results for the most fit solutions in the final populations across all 3-folds. The 80:20\% setting produced the highest sensitivity and specificity, but also used the highest number of features. As this was an unbalanced classification problem, there was high risk of Type II error in that heroin users were more likely to be incorrectly classified as Non-Users. For example, certain unfit models resulted in 100\% specificity and 0\% sensitivity, in which all Heroin-User were classified as Non-Users. As such, sensitivity, the correct identification of Heroin users, was more challenging to obtain than specificity. Sensitivity seemed to correlate with high weights for accuracy; however, the five data settings as in Table~\ref{tab:fold_avg}  were not enough to conduct a thorough statistical validation of that observation.

\begin{table}[htb]
\centering
\caption{\label{tab:fold_avg} Average Results Across 3-Folds.}
\small
\begin{tabular}{ccccc}
\hline
Group & Sensitivity & Specificity & Features & Fit  \\ 
\hline
A:80 F:20 & .725 & .934 & 7.156 & .588  \\ 
A:65 F:35 & .717 & .929 & 4.877 & .547 \\ 
A:50 F:50 & .707  & .927 & 3.244 & .534 \\ 
A:35 F:65 & .672  & .930 & 2.49 & .517 \\ 
A:20 F:80 & .458 & .952 & 1.62 & .499 \\  
\hline
\\
\end{tabular}
\end{table}

We conducted statistical significance tests of differences in sensitivity between weight conditions; see Table \ref{tab:sig}. We could not apply one-way ANOVA because the data did not meet the assumption of homogeneity of variance; therefore, we employed  multiple pairwise t-tests with no assumption of equal variances. We found that  
weight settings that favored accuracy were often significantly better at identifying heroin users. 

\begin{table}[htb]
\centering
\caption{\label{tab:sig} Sensitivity differences assessed with pairwise t-tests with non-pooled standard deviations. The lower the p-value, the higher is the significance of difference.}
\small
\begin{tabular}{ccccc}
\hline
 & A:80 F:20 & A:65 F:35 & A:50 F:50 & A:35 F:65  \\
\hline
A:65 F:35 & 0.375 & - & - & -  \\ 
A:50 F:50 & 0.165 & 0.402 & - & - \\ 
A:35 F:65 & .002**  & .007** & .065 & - \\ 
A:20 F:80 & $<$.001***  & $<$.001*** & $<$.001*** & $<$.001*** \\  
\hline
\\
\end{tabular}
\end{table}

\begin{figure}[htb]
\vskip 0.2in
\begin{center}
\textbf{Most Selected Features by Weight Setting}\par\medskip
\centerline{\includegraphics[height=6cm,width=1.1\columnwidth]{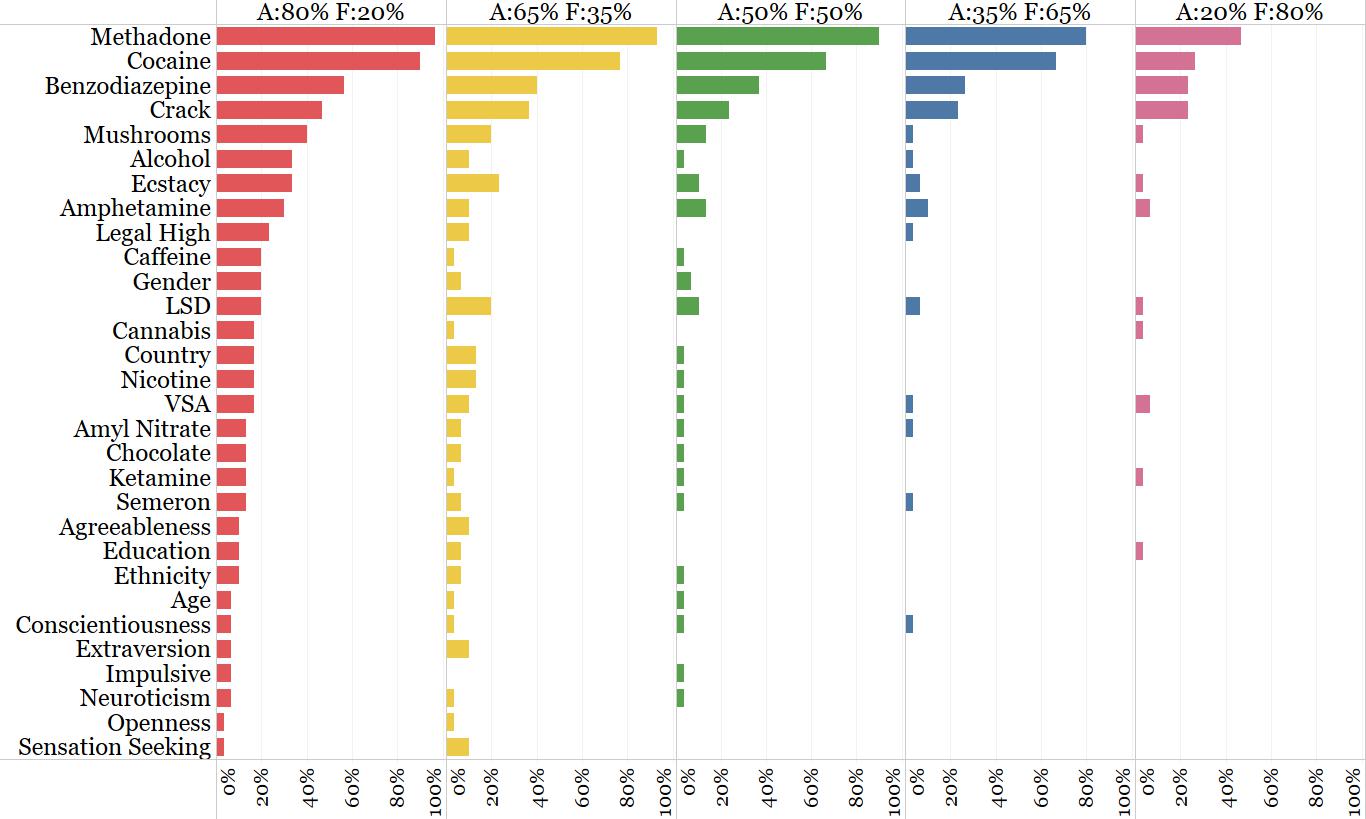}}
\caption{Features related to drug use were selected most often.}
\label{fig:selected}
\end{center}
\vskip -0.2in
\end{figure}

A key aspect of this research is to understand the major predictors of Heroin use; Figure \ref{fig:selected} plots the prevalence of various predictors in the classification models. The most selected features were consistent across all conditions, and these features all concern the use of other drugs. Therefore, consumption of Methadone, Cocaine, Benzodiazepines, and Crack were the most predictive features in classifying Heroin consumption. The presence of Benzodiazepines and Methadone is particularly revealing and troublesome because they are common prescription medications.

However, as expected, higher weights for the number of features discourage a high number of features, and they produced final models with a smaller number of features. While the 80:20\% combination used almost all features across generations, the 20:80\% condition rarely used more than the top four features. Overall, the demographic and personality variables are far less predictive than the drug related variables. Among these, Gender was selected most often.

\begin{table}[htb]
\centering
\caption{\label{tab:best} Most Accurate Heroin Classification Models.}
\small
\begin{tabular}{cccccc}
\hline
 & A:80 & A:65 & A:50 & A:35 & A:20 \\ 
 & F:20 & F:35 & F:50 & F:65 & F:80 \\\hline
Sensitivity & .814 & .771 & .814 & .771 & .771 \\ 
Specificity & .953 & .942 & .946 & .955 & .942 \\ 
Fit & .635 & .648 & .427 & .351 & .545 \\ 
Feature Reduce\% & 53.33 & 93.33 & 60.00 & 76.66 & 93.33 \\ 
SVM Cost & 87.83 & 58.56 & 32.26 & 36.62 & 42.70 \\ 
SVM Gamma & .253 & 2.97 & .312 & .583 & 3.16 \\ 
\hline
\\
\end{tabular}
\end{table}

The most predictive models, in terms of feature fitness and total accuracy, for each weight setting are shown in Tables \ref{tab:best} and \ref{tab:best_feat}. Each of these models have higher than average Sensitivity and Specificity for their condition. All of the models successfully reduced the number of features and converged on an accurate solution, however they are not all the most fit solutions in the set. For example, the 35\% 65\% model is highly accurate and reduces features, however it is less fit than average for its group. The SVM parameters selected also vary widely across conditions, and were not consistent with fitness weights. 

\begin{table}[htb]
\centering
\caption{\label{tab:best_feat} Features used by the most accurate Heroin classification models from each weight setting.}
\small
\begin{tabular}{cccccc}
\hline
Target  & A:80 & A:65 & A:50 & A:35 & A:20 \\ 
Feature & F:20 & F:35 & F:50 & F:65 & F:80 \\\hline
Alcohol & X &  &  &  &   \\ 
Amphet. & X & & X & X &   \\ 
Amyl. &  &  & X &  &   \\ 
Benzo. & X &  & X &  &   \\ 
Cannabis & X &  &  &  &   \\ 
Chocolate  &  &  &  &  &    \\ 
Cocaine & X & X & X & X & X   \\ 
Caffeine &  &  &  &  &   \\ 
Crack & X &  & X & X &   \\ 
Ecstasy &  &  & X & X &   \\ 
Ketamine &  &  &  &  &   \\ 
Legal Highs &  &  &  &  &   \\ 
LSD & X &  &  &  &   \\ 
Methadone & X & X & X & X & X  \\ 
MMushrooms & X &  & X & X &   \\ 
Nicotine &  &  &  &  &   \\ 
VSA & X &  & X &  &   \\ 
Semer & X &  &  &  &   \\ 
Age &  &  &  &  &  \\ 
Gender &  &  & X &  &  \\ 
Education &  &  &  &  &   \\ 
Country &  &  &  &  &   \\ 
Ethnicity & X &  &  &  &   \\ 
Agreeable. &  &  &  &  &   \\ 
Neuroticism &  &  & X &  &   \\ 
Extraversion & X &  &  &  &   \\ 
Oppenness &  &  &  &  &   \\ 
Consc. &  &  &  & X &   \\ 
Impulsivity & X &  & X &  &   \\ 
Sens. Seek &  &  &  &  &   \\   
\hline
\\
\end{tabular}
\end{table}

We compared these results with an SVM model that used all the features, with default $C$ and $\gamma$ parameters (1, 0.33). The average sensitivity for this model across the same 3-folds was 0.533 and average specificity was 0.988. As seen in the weight comparison, there was a high risk of Type II error, and the default model misclassified heroin users as non-users more often than the GA-SVM hybrid model.

\subsection{Discussion}
\label{sec:discussion}

This study accurately classified Heroin users by employing the GA-SVM Hybrid algorithm. Not only did the adopted method outperform standard benchmarks for classification such as SVM, the best model from this study produced greater overall accuracy (total of 176.7\%) than the previous best heroin prediction (total of 155.53\%) \cite{Fehr15} on this data set; for the purposes of this work \cite{Fehr15} is the baseline study. A key to this improvement is the addition of self-reported information about previous drug consumption in the data set for this research, unlike previous predictive studies. 

The understanding of prior drug use can be seen as privileged information as the accuracy of self-report drug use techniques have been put into question \cite{Garg15}. However previous research has linked consumption of Heroin with several types of drugs, and this information may be useful in identifying high-risk individuals. Research has shown that general Opioid use is correlated with the consumption of other drugs such as Cocaine, Cannabis, Hallicinogens \cite{Darke91}. In addition, the baseline study also indicated that Heroin use was correlated with consumption of Crack, Cocaine, and Methadone \cite{Fehr15}. In the larger context of the Opioid Epidemic, the current study showed that not only was Heroin consumption linked to illegal street drugs, but rather disturbingly also to common prescription medications such as Benzodiazepines and Methadone. Figure\ref{fig:selected} also shows that knowledge of the individuals previous drug consumption was more predictive than knowledge about their psycho-social factors as traditionally considered in previous studies. 

There has been a clear connection made between Heroin consumption and prescription medication in the USA over recent months, as it has been found that the company producing Opioid based Oxycontin knowingly pushed the drug onto high risk users \cite{Pitzke19}. These results further validate the relationship between consumption of Heroin and other drugs. However, the results may be limited by the fact that this study lacks information about whether users were prescribed these drugs, or if they were received illegally. 

Methodologically, these results showed that the choice of accuracy parameter and number of feature weights in the popular GA-SVM Hybrid fitness function have a significant impact on the outcome of the model. When the weights were adjusted, so did selected features, and values for SVM parameters. Subsequently, the researchers' choice of weights influences their final outcome and thus the importance of predictive features. As such, researchers should always report these weights or may consider a systematic approach to weight selection as shown in this study. 

\section{Conclusions}
\label{sec:conclusions}
Machine learning models have been applied to understand drug consumption patterns. These methods have the potential to support efforts to fight drug addiction; however they are sensitive to input features and model parameters. The GA-SVM Hybrid model improves upon previous research by automatically selecting optimal SVM parameters while simultaneously identifying the most predictive features. Though this is a powerful pairing, this study has shown that the choice of hyperparameters in GA has a substantial impact on model performance. Slight variations in the fitness function can significantly impact classification accuracy and the resulting importance of input features.

Due to the increase in Opioid related deaths, research has focused on the factors that may lead to Heroin consumption, such as the abuse of prescription pain medication. This research shows that Heroin consumption is related to consumption of other illegal drugs such as Cocaine. By including information about consumption of other drugs, and simultaneously selecting optimal model features and parameters, this study achieved a model that was more accurate in predicting Heroin use than previous research, while using a limited number of features. Results show that while both information about drug consumption and psycho-social factors support the identification of Heroin users, knowledge about previous drug consumption is significantly more predictive. These results may help health practitioners identify individuals at risk of heroin use.

  


\bibliographystyle{ACM-Reference-Format}
\bibliography{example_paper} 

\end{document}